\documentclass[11pt]{amsart}
\usepackage{graphicx}
\usepackage{amssymb,amscd}
\usepackage{array, amsmath, amssymb, amsthm, amsxtra, amscd, latexsym, graphics, psfrag, epsfig}
\usepackage{enumitem}

\def\url#1{\it #1\par}

\theoremstyle{plain}

\theoremstyle{remark}

\usepackage{prettyref}

\usepackage{algorithmic, algorithm}

\usepackage{arcs}

\begin{document}


%
%

\title{A two-step machine learning approach for crop disease detection: an application of GAN and UAV technology}
\renewcommand{\shorttitle}{A two-step machine learning approach for crop disease detection}

\author{Aaditya Prasad, Nikhil Mehta
}
\address{Tesla STEM High School\\
Redmond, WA, 98053, USA \\
Aaditya.prasad@outlook.com, shugis2op@gmail.com}

\author{Matthew Horak}

\address{Lockheed Martin Corporation \\
	Denver, CO, 80128, USA \\
matthew.horak@lmco.com}

\author{Wan D. Bae}

\address{Computer Science Department, Seattle University\\
	Seattle, WA, 98122, USA\\
	baew@seattleu.edu}

\maketitle

\begin{abstract}
Automated plant diagnosis is a technology that promises large increases in cost-efficiency for agriculture. However, multiple problems reduce the effectiveness of drones, including the inverse relationship between resolution and speed and the lack of adequate labeled training data. This paper presents a two- step machine learning approach that analyzes low-fidelity and high-fidelity images in sequence, preserving efficiency as well as accuracy. Two data-generators are also used to minimize class imbalance in the high-fidelity dataset and to produce low-fidelity data that is representative of UAV images. The analysis of applications and methods is conducted on a database of high-fidelity apple tree images which are corrupted with class imbalance. The application begins by generating high-fidelity data using generative networks and then uses this novel data alongside the original high-fidelity data to produce low-fidelity images. A machine-learning identifier identifies plants and labels them as potentially diseased or not. A machine learning classifier is then given the potentially diseased plant images and returns actual diagnoses for these plants. The results show an accuracy of 96.3\% for the high-fidelity system and a 75.5\% confidence level for our low-fidelity system. Our drone technology shows promising results in accuracy when compared to labor-based methods of diagnosis. 
\end{abstract}


\section{Introduction}	
By 2050, human agricultural crop yield will need to increase by an estimated 70 percent to sustain the expected population size. Crop diseases currently reduce the yield of the six most important food crops by 42 percent, and some farms are wiped out entirely on an annual basis\cite{faoreport}. The first step that needs to be taken to start the extermination of crop pests and diseases is a method for accurate diagnosis.  However, current methods to obtain these diagnoses are heavily limited in both scale and speed. These methods involve the hiring of laborers on the ground to simply walk the fields and look for signs of pestilence such as discolorations or damage. When potentially sick plants are found, a reliable diagnosis requires either another hired expert on hand or for the plant to be sent to a nearby university or lab, which is very time-consuming and expensive.

In recent years, technology has arisen with the capability to diagnose plants with a simple image or reading taken by a smartphone; yet still, this requires the hiring of a human on the ground to walk around and take pictures. For many farms, which in many regions of the US commonly exceed 400 acres in size, this requirement makes accurate and timely diagnoses hard to obtain. In the last few years, attention has turned to the use of Unmanned Aerial Vehicles (UAVs) paired with large-scale backend systems involving machine learning technologies to complete this task. While UAVs offer a faster automatic detection of diseased plants, they present two major challenges that will have to be addressed before commercial use is viable \cite{barbedo2019review}.

The first challenge is that the image classification problem, required for disease diagnosis from plant images, is exceedingly difficult on UAV images for several reasons.  First is the low resolution of images taken by a UAV flying high above a field.  Second is the extreme variability of crop images in the field in terms of lighting, angle, soil color and atmospheric conditions.  Third is that crop diseases range in scope from affecting all the plants in an entire area to affecting parts of individual plants. Finally, UAV images of crops almost always contain background content of the ground or nearby plants that must be removed from the analysis.

The second problem specific to plant disease detection is the lack of high quality, labeled data. Solid data is the backbone of any machine learning classifier, and the availability of a large and varied training data set would go a long way to solving the first challenge mentioned above.  The problem is that up to now, producing labeled UAV images of diseased plants involves the time-consuming task of manual disease detection mentioned above coupled with UAV-based imaging of the same plants.  Therefore, producing sufficient labeled data by hand to adequately train image classification systems on UAV images is counterproductive in terms of shorter-term benefits for farmers and other stakeholders in a position to fund and benefit from the longer term from improvements in UAV-based detection technology.

In this paper, we propose a machine learning framework consisting of two integrated components, data generating and modeling.  Each of these components includes two main steps.  For the modeling component, we propose a standard two step identification and classification scheme.  In the identification step, the model identifies potentially diseased regions of a field in real time from low-resolution UAV images.  For the classification step, the UAV navigates closer to the potentially diseased regions and captures high resolution images for disease classification.  This two-step model facilitates our novel data generation scheme that couples a generative adversarial network with a new low fidelity data generator.  The adversarial network produces high quality synthetic training images for the classifier and the low fidelity generator assembles the resulting high resolution images into synthetic low resolution images of crop fields for training the identifier.  In this way, we are able to generate sufficient synthetic data to train both steps of the model from a single small public-domain data set of high resolution images of crop diseases.  Once the two step model is trained and validated using the above framework, it is used in its two-step identification/classification form.

This paper has two main contributions.  First is the way that we use the identification/classification architecture of the model to facilitate synthetic data generation on two resolution scales.  The second is the technical method for generating low resolution synthetic crop field images used in the low fidelity data generator.  Our framework is agnostic to the particular Artificial Intelligence (AI) and Machine Learning (ML) tools used to perform the identification, classification and high resolution image generation.  Therefore, as these core AI and ML tool continue to improve over time, the gains will translate directly to our framework as well.  Finally, we remark that our solution is not limited to crop disease identification but can be applied to any classification problem involving low and high resolution data, especially those common in geospatial disciplines.

The rest of the paper is organized as follows. In Section 2, we discuss related work. Section 3 presents our two component machine learning framework for automatic crop disease diagnosis and provided a detailed description of our framework.  Section 4 is a summary of the tools and data sets used in our experiments and a demonstration of the framework, including the AI/ML models, hardware, and performance on the test set.  We conclude with directions for future work in Section 5.

\section{Related Work}
\label{sec:related}
Research associated with automatic crop disease detection using imagery data has received considerable attention. Clive et al. \cite{bock2020visual} present a literature survey of automatic classification studies. Accuracy of these studies generally range from 90\% to 98\%. On the other hand, another survey paper \cite{del2017standard} indicates that on a variety of tasks, human accuracy averages around 95\% to 97\%, with significant person to person variability. Clive et al. also point out that rapid loss of expertise and ability is another difficulty that plagues manual diagnosis of crop diseases.

Particular interest has been paid to disease detection by UAVs.  Recent study attempts to estimate the fraction of a potato field infected with blight from UAV images \cite{duarte2018evaluating}.  Their models achieve maximal coefficient of determination $r^2$ of 0.75.  Similarly, Sugiura et al. \cite{sugiura2016field} report coefficient of determination of 0.77 at the slightly different task of estimating the overall disease severity in potato fields from UAV imagery.  More generally, several recent papers \cite{barbedo2019review,singh2012taro,gao2020framework} present the current state of automated crop diagnosis by UAVs.  They indicate that the current techniques for automated disease detection are often limited in their scope and dependent on unpractical assumptions. This lack of advancements is from various problems: the presence of complex backgrounds that cannot be easily separated from the area of interest; boundaries of the symptoms often are not well defined; uncontrolled capture conditions may present characteristics that make the image analysis more difficult, and so on.  Several specific challenges in detecting crop diseases identified in these works are: (1) the lack of labeled data, (2) model accuracy, and (3) system efficiency. This motivates us to develop our two-step machine learning approach that separately analyzes low-fidelity and high-fidelity images and the use of GANs. 



Our framework was built upon a system information of agriculture presented \cite{fasoula2012nonstop,thapa2020plant,hughes2015open}. Fasoula et al. \cite{fasoula2012nonstop} propose selection criteria based on the genetic and epigenetic responses of healthy and superior crops. The work in research \cite{thapa2020plant,hughes2015open} contributes toward development and deployment of machine learning-based automated plant disease detection by providing repositories of plant images. Thapa et al. \cite{thapa2020plant} manually collected 3,651 high-quality, real-life symptom images of multiple apple foliar diseases, with variable illumination, angles, surfaces, and noise. They also made a synthetic dataset created for apple scab, cedar apple rust, and healthy leaves available to the public. Hughes et al. \cite{hughes2015open} shared over 50,000 expertly curated images on healthy and infected leaves of crop plants through an existing online platform PlantVillage. Our experiments were conducted on the Plant Pathology 2020 datasets \cite{thapa2020plant}.

Nazki et al. \cite{nazki2020unsupervised} discuss the common problems caused by lack of data in the task of diagnosing tomato tree diseases and proposed the use of GANs to alleviate class imbalance. They showed that the GANs improved the model accuracy when synthetic images were appended to the dataset of a standard CNN. The authors also presented AR-GAN, a novel method of reducing mode collapse by optimizing on an Activation Reconstruction loss. This work gave us concrete evidence that synthetic data from GANs can be used to improve CNN results, and also led us to DCGANs, which were suitable for our framework. Although the core ideas are similar, the specific architecture of our DCGAN differs from the one they proposed.

Several base models of Generative Adversarial Networks (GAN) and their implementations were presented in several recent papers \cite{joshic2020,goodfellow2020generative}. Generative Adversarial Networks are pairs of neural nets that are split into two roles: generator and discriminator. The generator learns to develop synthetic images of some class, while the discriminator learns to discern between real and synthetic images. The models train off of each other to improve results. Further, recent work \cite{mirza2014conditional}  proposed a specific archetype of GAN called a conditional GAN, which has a slightly different use case. We conducted extended experiments on the models proposed by other researchers \cite{joshic2020,goodfellow2020generative,mirza2014conditional} with all of those structures to learn from the use cases and troubleshooting strategies proposed in these papers. We then propose a new GAN-based structure that is more suitable for the crop disease detection modeling.

Recent advances in machine learning have provided new methods and technologies for classification modeling and object detection. Wang et al. \cite{wang2017automatic} present a series of deep convolutional neural networks  trained to diagnose the severity of the apple black rot disease using datasets in the PlantVillage \cite{thapa2020plant}, which are annotated by botanists with four severity stages. Their best classifier model yields an overall accuracy of 90.4\%. The identifier and classifier in our work were developed based upon the methods and procedures proposed in several recent papers \cite{kingma2013auto,perez2017effectiveness,simonyan2014very}. In particular, we adopted various augmentations suggested by Perez et al. \cite{perez2017effectiveness} in our work.

\section{Two-Step Machine Learning Framework}
\label{sec:frame}
\subsection{Data Generation and Modeling Framework}
We start with a small public domain data set of high resolution images of diseased plants from the Plant Pathology dataset.  This small data set of images is augmented with the High Fidelity Datagen, which is an adversarial network, described in \prettyref{sec:HFDG}.  Once sufficiently many high resolution images are produced, they are passed to the Low-Fidelity Datagen, described in \prettyref{sec:LFDG} for generation of synthetic images of crop fields.  The identifier component of the diagnosis model is an image segmentation algorithm that is trained on the low resolution crop images to isolate potentially diseased regions of the image.  The classifier component is a standard multi-class image classifier that identifies the disease present in a high resolution image of a diseased plant. The proposed framework is illustrated in Figure 1.

\begin{figure}[th]
	
	\centerline{\includegraphics[width=12cm]{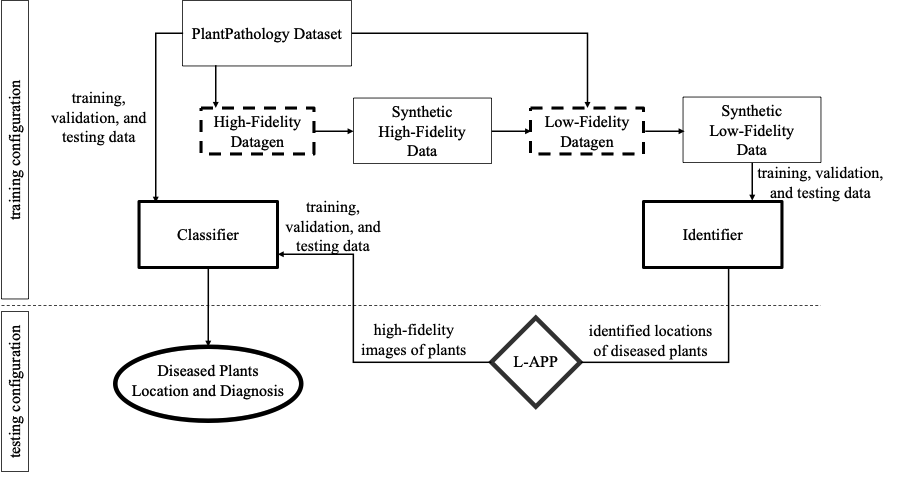}}
	\vspace*{8pt}
	\label{fig:fig1}
	\caption{An overview of data generation and modeling framework.}
\end{figure}

\subsection{Modeling Pipeline: Identifier and Classifier}
Once the identifier and classifier are trained, these two models are integrated into a two-step ML application as deployed in the field. We now present this full model’s usage as illustrated in Figure 2.  First, using the proper hardware interface, the drone is operated high above the field to capture low resolution images of the field.  In real time, potentially diseased regions of these images are identified and the drone is maneuvered near to the potentially diseased plants to capture high resolution images of the potentially diseased plants that are passed to the classifier for diagnosis.  Since both of the models we use for the image segmentation and image classification tasks of the model are industry standard, we postpone their brief discussion to \prettyref{sec:vision}. The main innovation of this paper is in the way we harness the division of labor of the segmentation and classification tasks to facilitate generation of sufficiently many labeled images for training. 

\begin{figure}[th]
	
	\centerline{\includegraphics[width=12cm]{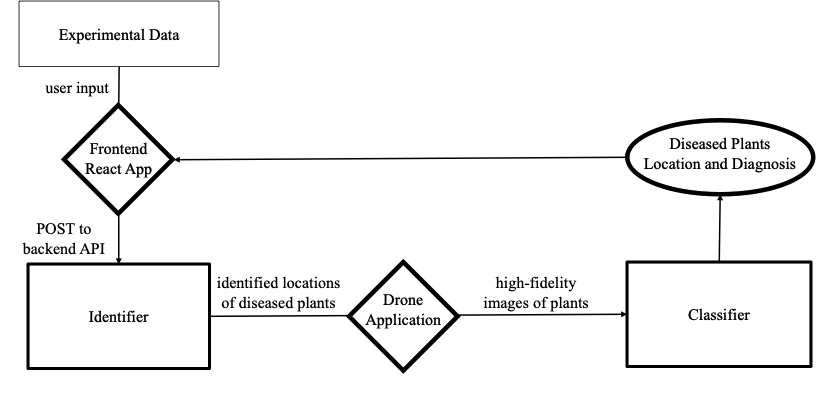}}
	\vspace*{8pt}
	\label{fig:fig2}
	\caption{The pipeline with identifier and classifier.}
\end{figure}


\subsection{Data Generation}
\subsubsection{High Fidelity Data Generation}
\label{sec:HFD}
Since the classification task is performed strictly on high resolution images of diseased plants, training data for this step is simply close-up high resolution images of diseased plants together with the disease label. The Plant Pathology dataset contains a modest number of such images, but not enough to train the classifier.  In addition to the data set being insufficiently large to train the classifier, it is also plagued by class imbalance, as discussed in \prettyref{sec:datasets}. We solve both the data scarcity and class imbalance by augmenting the data set with a Deep Convolutional Generative Adversarial Network (DCGAN).  Again, the particular DCGAN instance we use is industry standard, so we postpone a brief discussion to \prettyref{sec:HFDG}. Because GANs are usually configured to produce images for a single label, multiple GAN’s were trained so that we could produce novel images of each label.

\subsubsection{Low Fidelity Data Generation}
\label{sec:LFD}
The key insight that allowed the full identification and classification model to be trained on a modest number of high resolution close-up images is that the high/low resolution division of labor in the identification/classification steps of the model allow us to take advantage of the high quality synthetic close up plant images to generate low resolution far field images of whole fields.  We developed our new Low Fidelity Datagen (LFD) for this purpose, which we describe in detail in \prettyref{sec:LFDG}

%

\section{System Performance and Evaluation}
In this section, we describe our source of training and testing data and the hardware and software tools we use.  Recall from \prettyref{sec:frame}, that we will pass data images from our date source, Plant Pathology 2020 \cite{thapa2020plant}, through the High-Fidelity Datagen to create synthetic high-fidelity data, which is then used alongside the original dataset as data for the Low-Fidelity Datagen. The synthetic low-fidelity data produced by the Low-Fidelity Datagen is passed to the Identifier. The Plant Pathology 2020 real datasets as well as synthetic high-fidelity data is also passed to the classifier to train on. For testing purposes, specific high-fidelity images from the real datasets were marked as the test set. This test data was used to evaluate performance of the model components individually. The results from the individual models are used to estimate the performance of the entire end-to-end model pipeline.

\subsection{A Tech Stack Overview}
\label{sec:stack}
A DJI Marvic 2 Pro drone, one of the most popular drones, was used for validating the proposed system framework. This drone is known to provide high camera quality and the best obstacle detection among DJI models. Note that our framework can work as-is with most DJI models with minor modifications using the DJI API.

A tech stack is the combination of technologies a company uses to build and run an application or project. Sometimes called a “solutions stack,” a tech stack typically consists of programming languages, frameworks, a database, front-end tools, back-end tools, and applications connected via APIs.

The tech stack used for our system focuses on agility of development that improves flexibility, balance, and control in the system. Hence, it helps to maintain and update the system. Three main technologies in our tech stack are: (1) backend tools, (2) frontend tools, and (3) applications. The front-end tools include an Android application and a website using the popular state-management framework React.js. As part of managing HTTP requests to the backend API, the system uses  the Axios library for the React portion of the frontend, and the built-in Volley library for the Android application. Django, a convenient ORM tool, is used for managing the backend SQLite database/endpoint registry. Django Rest Framework provides an environment in which we can easily create REST APIs. To fully implement Django, a Web server is hosted using Gunicorn through a WSGI entry-point on Heroku. Finally, the library Chakra is used to configure the UI of our React application. The following subsections describe the three main technologies.

The backend tools include three parts: (a) the machine learning applications (i.e., model weight files), (b) the Django REST API using Django Rest Framework (which allows our frontend to send HTTP requests to our machine learning applications), and (c) the Django backend itself (which handles the running of the machine learning applications in accordance with the API). Our API consists of a root which contains four subdirectories: one for a list of API endpoint URLs that can be used for an API view that uses the model, one for a list of ML algorithms available from the API, one which details the status of all current ML algorithms, and one that stores the requests for ML algorithms. To serve the intended functionality of our models through the API, the system uses endpoints for both the identifier and classifier. These predict views can be accessed by navigating to a REST API, which allows for POST requests and will return a response for probabilities.

The React Application on the frontend is relatively simple. It serves as a tool that makes HTTP  requests easy in a UI-focused manner. The tool consists of tabs for standard HTTP request methods including GET, POST, PATCH, PUT, and DELETE. This allows us to test our backend API with ease and additionally allows us to upload several images to our backend at once, which is especially important to generate results for our experimental data.

The Android DJI Mobile SDK is used for the mobile application development of controlling the DJI Mavic Pro 2 drone. Using the SDK and the Android HTTP request library Volley, the drone flies over a field while sending HTTP requests for each image frame to the Identifier (a backend model that predicts where diseased plants lie in a field) and using the API response to guide the drone. Once the drone is in the correct location to take a photo of the plant, the frame of the image will be sent to the classifier, which then returns a diagnosis of the plant to the user.

The system also utilizes two software tools for synthetic data generation. The Low Fidelity Datagen (LFD) is a generator which builds images that represent the pictures a UAV might take while flying over a field. These are low fidelity, high field of view (FOV) images. The High Fidelity Datagen (HFD) is a generator that produces new pictures of plants. These will henceforth be referred to as novel images, as opposed to the original images that are found in the Plant Pathology dataset \cite{thapa2020plant}. The generation method used is a DCGAN, or Deep Convolutional Generative Adversarial Network. More technical details of each of these generators will be discussed in \prettyref{sec:generators}.

\subsection{Datasets}
\label{sec:datasets}
The dataset is used in this paper is the Plant Pathology 2020 - FGVC7 dataset \cite{thapa2020plant}. It contains 1821 labeled images of plants.  The dataset also possesses labels for all these images in csv format. It is important to note is that the dataset exhibits a class imbalance. There are 416, 592, 622, and 91 images in the classes healthy, scab, rust, and multiple diseased, respectively. This lack of multiple diseased images as well as healthy images reduces the effectiveness of machine learning mode which are trained on these classed.

\subsection{Computer Vision Tools}
\label{sec:vision}
For the image classification task, which classifies into the disease categories high resolution images of the potentially diseased regions of the crop fields, we use the EfficientNet architecture, which is an industry standard high performing image classification nertwork.  We used Genetic Efficient Nets for PyTorch \cite{rwightman}, which specializes in providing a range of pre-trained EfficientNet and MobileNet models in the required PyTorch format. This EfficientNet uses a baseline network that was created through a neural architecture search (i.e., an automated architecture engineering scheme) to optimize accuracy and FLOPs.
The implementation of this architecture is found at a GitHub repository \cite{rwightman}.

For the image segmentation task of isolating potentially diseased regions of the crop field, we use the  EfficientDet, network, which is an industry high performing model used for object detection.  We selected EfficientDet for the same reason as we selected EfficientNet:  EfficientDet outcompetes other models due to its remarkable scaling ability and FLOP reduction

Both EfficientNet and EfficientDet achieve their efficiency using the same compound scaling method that uniformly scales the depth, resolution, and width for the backbone and network at the same time.  EfficientDet also utilizes EfficinetNet backbones due to their efficiency and scalability alongside a weighted bi-directional feature pyramid network (BiFPN) that creates weights while applying feature fusion. This bi-directional pyramid network is intended to solve the problem of other previously used networks which sum features up without distinction because input features at different resolutions contribute to the output feature unequally. 

\subsection{Data Generators}
\label{sec:generators}
\subsubsection{High Fidelity Data Generator}
\label{sec:HFDG}
We use an industry standard Generative Adversarial Network (GAN) to generate additional synthetic high resolution images of plants and plant diseases.  In essence, a GAN works by pitting two networks, a generator and a discriminator, against each other. The generator attempts to create images that will fool the discriminator, while the discriminator attempts to classify images as real or generated.
The DCGAN works analogously to a general GAN, with the added clarification that it specifically uses convolution and convolutional-transpose layers in the discriminator and generator, respectively. Analysis was not done on differing levels of effectiveness between various GAN archetypes, but there is no reason to believe another configuration would lead to a much better output or training time than the one used in this paper.  We made no changes to the architecture from the original paper by Radford et. al. \cite{radford2015unsupervised}

\subsubsection{Low Fidelity Data Generator}
\label{sec:LFDG}
As mentioned in \prettyref{sec:LFD}, the LFD is a generator which builds images that represent the pictures a UAV might take while flying over a field by assembling both real and synthetic high-fidelity data images into high field of view images of mock ``fields". This low-fidelity synthetic data will then be used to increase the amount of data available to train, validate, and test the model performing the image segmentation task.  For each high field of view image, The LFD also outputs a csv that contains the labels of the locations in the synthetic image of the diseased areas.  An example image-csv pair is shown in Figure 3 and Table 1 and described in detail below.

\begin{figure}[th]
	
	\centerline{\includegraphics[width=6.5cm]{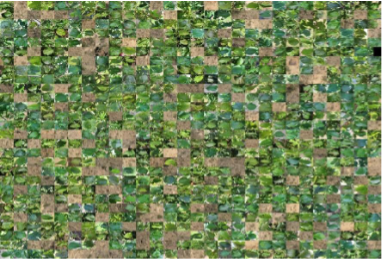}}
	\vspace*{8pt}
	\label{fig:fig3}
	\vspace{-3mm}
	\caption{Example image output from the LFD.}
\end{figure}

\begin{table}[tpbh]
	\vspace{5mm}
	\label{tab:LFD}
	\caption{Example csv format output from the LFD.} 
		\begin{tabular}{lcc} 
			\hline
			id &  bbox & class label  \\
			\hline
			Train\_1609.jpg \hspace{5mm} &  [64, 0, 64, 43]&	1 \\
			\hline
			Train\_1028.jpg&  [128, 0, 64, 43]&	1 \\
			\hline
			Train\_354.jpg&  [192, 0, 64, 43]&	1 \\
			\hline
			Train\_1082.jpg&  [256, 0, 64, 43]&	0 \\
			\hline
			Train\_10.jpg&  [320, 0, 64, 43]&	1 \\
			\hline
			Train\_1280.jpg&  [384, 0, 64, 43]&	1 \\
			\hline
			Train\_463.jpg&  [448, 0, 64, 43]&	1 \\
			\hline
			Train\_1178.jpg&  [512, 0, 64, 43]&	1 \\
			\hline            
	\end{tabular}
\end{table}

An output image from the LFD is a picture of size 1792 x 1204 pixels. It is divided into 28 rows and columns, with each position holding a 64 x 43-pixel picture of either plant part of dirt.  In our test case, the plant was always a set of apple leaves because our experimental tests were performed on apple orchards.  Moreover, the dimensions of the output as well as the subimages was selected to be representative of the hardware in our experimental tests, which was a DJI Mavic 2 Pro 20 feet above the apple canopies, described in \prettyref{sec:stack}.

The subimages that make up an output image are mostly plants, but there are images of soil. These were added to force the identifier to learn how to differentiate foliage with background soil, which are the two main features in an aerial image of an apple orchard. The specific sub-image used for the soil is a random slice of a large soil texture. The soils are randomly scattered throughout the images at an approximately 1:5 ratio with images of leaves.

Second, each sub-image which contains a plant has a corresponding row in the csv for the overall image. The first column is the id, which is simply the name of the original image. This allows other applications to find and use those original, full size images. The second column is the bounding box. This is the location of the subimage represented by that row in the overall image. Finally, the last column is “sick”. These plant images come from the Plant Pathology dataset \cite{thapa2020plant}, and are labeled as ``healthy", ``rust", ``scab", or ``multiple diseases". The last three indicate by a ``1" in the corresponding sick column, while a ``healthy" image receives a 0.  Rows are not made for soil images.

After the synthetic data has been processed and augmented into usable low fidelity training images, we then augment the low fidelity dataset by performing operations on images known as cutout and CutMix, which generate new training samples from old by combining two training samples. While we augment data using a both algorithms, CutMix serves a more important role as it makes use of the entire image area, and mixes labels so the model can learn to differentiate even within images. We use CutMix on a random basis – for every item in the dataset, there is a 50\% chance for the Dataset class to load a CutMixed image rather than a normal one.

\subsection{Model Training and Testing}
The system will be trained on a set of diseases that afflict Apple Trees. The specific diseases are rust and scab, and there are also healthy and multiple diseased pictures as well.  Since the system is essentially a set of machine learning models, it can be used on a wide range of plants/diseases with little to no modification if data for those plants and diseases is easily accessible and of a high quality. We chose apple trees but our system should be trainable on other plants and diseases if resources for those plants/diseases become available.

We trained our implementation of EfficientDet, on the loss function known as summary loss because the summary loss requires optimization both the difference between bounding box outputs with their targets (i.e., a box regression loss) and class outputs with class targets (i.e., a cross-entropy measurement of the classification for each box). An example image-csv pair is shown in Figure 3 and Table 1. Table 2 shows an example of a slice of training data in csv format. 


\begin{table}[tpbh]
	\vspace{5mm}
	\label{tab:slice}
	\begin{center}
	    \caption{A slice of training data in csv format} 
		\begin{tabular}{lcccc} 
			\hline
			image\_id &  \# healthy & \# multiple\_diseases & \# rust & \# scab \\ \hline
			Train\_0.jpg&  0 & 0 & 0 & 1 \\
			\hline
			Train\_1.jpg&  0 & 1 & 0 & 0 \\
			\hline
			Train\_2.jpg&  1 & 0 & 0 &  0 \\
			\hline
			Train\_3.jpg&  0 & 0 & 1 & 0 \\
			\hline
			Train\_4.jpg& 1 & 0 & 0 & 0 \\
			\hline
			Train\_5.jpg&  1 & 0 & 0 & 0 \\
			\hline
			Train\_6.jpg&  0 & 1 & 0 & 0 \\
			\hline
			Train\_7.jpg&  0 & 0 & 0 &	1 \\
			\hline     
			Train\_8.jpg&  0 & 0 & 0 & 1 \\
			\hline
			Train\_9.jpg&  1 & 0 & 0 & 0 \\
			\hline
			Train\_10.jpg&  0 & 0 & 1 &	0 \\
			\hline                
		\end{tabular} 
	\end{center}
\end{table}

With a fully trained EfficientDet, trained on synthetic low-fidelity data, we use two additional algorithms to further push the prediction estimates of the identifier. These algorithms are Weighted Boxes Fusion (WBF) alongside Test-Time Augmentation (TTA). The function of the TTA approach is simple – while testing the classifier, we perform augmentations (such as flipping the image or rotating it), which mimic the original augmentations performed and then average the predictive value. For the present application this provided a performance boost of approximately 5 percentage points. Additionally, TTA replaces the need for an ensemble consisting of many models, as it provides several predictions which can be used with the Weighted Boxes Fusion algorithm we use. In our approach, Weighted Boxes Fusion serves as an algorithm to replace the current popular method of merging ensemble boxes, Non-maximum Suppression (NMS). In the NMS approach, boxes are considered as belonging to one object if their overlap is higher than a threshold. The issue with this is that it completely disregards the non-overlap predictions that correspond to the target. However, this changes with a Weighted Boxes Fusion approach: with WBF, the correct box is picked, but then additional parameters from other semi-accurate boxes are applied to further correct the box, as shown in Figure 4.

\begin{figure}[th]
	
	\centerline{\includegraphics[width=7cm]{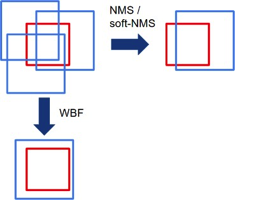}}
	\vspace*{8pt}
	\label{fig:fig4}
	\caption{A demonstration of outcomes from NMS and WBF approaches given TTA predictions (as a replacement for ensemble predictions). Red represents ground truth, and blue represents TTA predictions.}
\end{figure}

The performance of the identifier on unseen test data was measure by three statistics, positive confidence level on test data, negative confidence level on test data and accuracy on test data. The performance of the trained identifier is summarized in Table 3. 

\begin{table}[tpbh]
	\label{tab:identifier}
	\caption{Identifier summarized results.}
     \vspace{2mm}
		\begin{tabular}{cccc} 
			\hline
			Test  &  Average positive   & Average negative  & Accuracy \\ 
			summary loss & confidence on test data & confidence on test data & on test data \\ 
			\hline

			1.89142 & 0.54998 & 0.20468 & 0.75466 \\
			\hline            
		\end{tabular}
\end{table}

The Classifier is the model that handles the high-fidelity dataset. The goal of the classifier is to return a diagnosis of a plant into one of the following categories: healthy, multiple diseases, scab, or rust. The Classifier utilizes the machine learning framework TensorFlow running on a TPU with data stored in Google Cloud Storage (GCS) buckets. The classifier is much simpler than the identifier.

Data processing starts by decoding images to bits and shuffling them into TensorFlow TFRecordDataset objects. Data augmentation is then applied but is much simpler and only includes image flips. To regulate learning rate, we created a learning rate function which is managed by TensorFlow’s LearningRateScheduler class. We define the epochs for which learning rate ramps, sustains, and exponentially decays, as shown in Figure 5.

\begin{figure}[th]
	
	\centerline{\includegraphics[width=9cm]{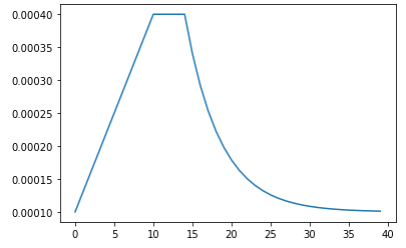}}
	\vspace*{8pt}
	\label{fig:fig5}
	\caption{Epoch (x-axis) versus learning rate (y-axis)}
\end{figure}

To define the model, we use TensorFlow’s implementation of EfficientNetB7, as discussed earlier. We compile the model using TensorFlow.Keras.Sequential, adding the EfficientNet and an additional final dense Softmax layer (with 4 units).  The optimizer we used was the Adam optimizer, which is a stochastic gradient descent method, simply because it is a very robust optimizer that historically performs well in a wide variety of classification problems.  Its robustness seems to stem mainly from its use adaptive learning rates for different parameters, which allowed us to focus more on tuning our architecture and less on particular hyperparameter settings such as learning schedules. Because we are using the Softmax  activation function, categorical cross entropy was our loss function.

A summary of the performance of the classification component of the model is shown in the confusion matrix for the classifier on unseen test data in Table 4.  The performance of the classifier on was measure by three statistics, Precision, Recall and F1-Score on unseen test data.  The performance of the final trained identifier is sown in Table 5. 

Overall, the average accuracy of our classification model was 96.3\% and the accuracy of our identifier model was 75.5\%.  If the performances were independent of each other, the probability of the full pipeline making a correct diagnosis would be the product of the probabilities of the two  individual models, which suggests a full pipeline accuracy of approximately 72.3\%.  On the other hand, if the samples that the classifier got wrong were disjoint from the samples that the identifier got wrong, the probability of an incorrect diagnosis would be the sum of the probabilities of each step giving an incorrect answer, which suggests a lower bound of 71.8\% for the accuracy of the full pipeline.  Finally, if the set of the classifier's errors is contained in the set of errors of the identifier, then the full pipeline accuracy would be equal to that of the identifier, suggesting an upper bound on the full pipeline accuracy of 75.5\%.

\begin{table}[tpbh]
	\label{tab:res1}
	\caption{Confusion matrix on test data for the classifier before GAN data is added.}
		\vspace{2mm}
		\begin{tabular}{lcccc} 
			\hline
		    &  Healthy (p) & Multiple diseases (p) & Rust (p) & Scab (p) \\ 
		    \hline
			Health (a) & 47 & 1 & 1 & 0 \\
			Multiple diseases (a) & 0 & 4 & 0 & 1 \\
			Rust (a) & 0 & 2& 55 & 0 \\
			Scab (a) & 0 & 1 & 0 & 52 \\
			\hline            
	\end{tabular}
\vspace{2mm}
\\ Note: A (p) label indicates the model’s predicted class while a (a) label indicates the actual class of the image.

\end{table}

\begin{table}[tpbh]
	\label{tab:res2}
	\caption{Classifier performance metrics}
		\begin{tabular}{lcccc} \\
			\hline
	&   Precision & Recall & $F_1$-score & Support \\ 
	\hline
	Health  & 0.96&1.0 & 0.98 & 47 \\
	Multiple diseases& 0.8 & 0.5 & 0.62 & 8 \\
	Rust  & 0.96 & 0.98 & 0.97 & 56 \\
	Scab & 0.98 & 0.98 & 0.98 & 53\\
	\hline            
	\end{tabular}
\vspace{2mm}
\\Note: Support shows the number of samples predicted to be that class within the test set and can be used as an indicator for the accuracy of other metrics.
\end{table}


\section{Conclusion and Future Steps}

At over $70\%$, the accuracy of our system seems promising, though ideally one might hope for higher accuracy.  This accuracy compares favorably to the typical $R^2$ values of approximately 0.7 reported in the studies of UAV enabled potato disease detection \cite{duarte2018evaluating,sugiura2016field}. We remark that although the  classification accuracy metric and the regression coefficient of determination are not directly comparable, all of these values indicate room for improvement. We also remark that the accuracy of our classifier model compares well to the current state of the art reported in \prettyref{sec:related} of 95\% to 98\% on similar tasks. 

Furthermore, our framework has two advantages over models common in the literature.  First, most of the UAV based models report classifications or disease severity ratings on the basis of the entire field, whereas our model reports specifiic locations of potentially diseased plants.  Second, our models are able to train to levels competitive with state-of-the-art models using only a small amount of labeled training data due to our novel data generation algorithms.

For improving our models' performance in the future, it is clear that the main hindrance to accuracy comes from the Identifier.  This might be expected because object detection is a much more complicated task than simple n-ary classification, resulting in generally low accuracies even with state-of-the-art models.  Therefore, the next future step for the improvement of the model accuracy will be to improve the accuracy of the identifier portion of the model.  Several approaches could be used here.  The first, is the address the possibility that the model is catering poorly to the data generated by the low fidelity datagen. Some ways we could accustom the model for this dataset are (1) more accurately labelling boxes around images so that they don't overlap, (2) recognizing the lack of padding within the data, which we could fix by predicting smaller boxes, and (3) using the Classifier to confirm or deny bounding box predictions within the Identifier. The second big hindrance to the Identifier was simply the size of the data. Even optimizing with a GPU, it took around 16 hours to run through around 2000 images. The problem with this is that it meant we were unable iterate through higher lever optimization to the model including identifying optimal hyperparameters and arrangements of the data augmentation steps.  This would have to be addressed through rigorous code optimization because run time issues are common throughout much of the identifier's code, especially in places that deal with data.


Finally, tests in this paper were conducted on only one crop, apples.  For a more complete evaluation of the gains achievable by our framework, it would have to be trained on tested on other major crops.  Finally, in the present paper our far-field, low resolution test data was artificially constructed form real images of diseased and healthy plants.  The next step in the direction of obtaining an estimate of the model's performance in the field with true drone data would be first to collect both far-field low fidelity drone images of crops and close-up high fidelity images of both healthy and diseased plants in the low fidelity drone images.  A small set of such images could be augmented to a larger training data set to train a field model, which would then be tested on a smaller hold-out set of the far-field and associated high fidelity images.  Admittedly, this process would require investment of time and money to generate labeled data, but as demonstrated in this paper, the data requirements of our model are very modest, which would keep training costs minimal.






\bibliographystyle{amsplain}
\bibliography{sample}

\providecommand{\bysame}{\leavevmode\hbox to3em{\hrulefill}\thinspace}
\providecommand{\MR}{\relax\ifhmode\unskip\space\fi MR }
\providecommand{\MRhref}[2]{%
  \href{http://www.ams.org/mathscinet-getitem?mr=#1}{#2}
}
\providecommand{\href}[2]{#2}
\begin{thebibliography}{10}

\bibitem{barbedo2019review}
Jayme Garcia~Arnal Barbedo, \emph{A review on the use of unmanned aerial
  vehicles and imaging sensors for monitoring and assessing plant stresses},
  Drones \textbf{3} (2019), no.~2, 40.

\bibitem{bock2020visual}
Clive~H Bock, Jayme~GA Barbedo, Emerson~M Del~Ponte, David Bohnenkamp, and
  Anne-Katrin Mahlein, \emph{From visual estimates to fully automated
  sensor-based measurements of plant disease severity: status and challenges
  for improving accuracy}, Phytopathology Research \textbf{2} (2020), no.~1,
  1--30.

\bibitem{del2017standard}
Emerson~M Del~Ponte, Sarah~J Pethybridge, Clive~H Bock, Sami~J Michereff,
  Franklin~J Machado, and Pi{\'e}rri Spolti, \emph{Standard area diagrams for
  aiding severity estimation: scientometrics, pathosystems, and methodological
  trends in the last 25 years}, Phytopathology \textbf{107} (2017), no.~10,
  1161--1174.

\bibitem{duarte2018evaluating}
Julio~M Duarte-Carvajalino, Diego~F Alzate, Andr{\'e}s~A Ramirez, Juan~D
  Santa-Sepulveda, Alexandra~E Fajardo-Rojas, and Mauricio Soto-Su{\'a}rez,
  \emph{Evaluating late blight severity in potato crops using unmanned aerial
  vehicles and machine learning algorithms}, Remote Sensing \textbf{10} (2018),
  no.~10, 1513.

\bibitem{faoreport}
FAO, \emph{Food and agriculture organization of the united nations:
  International plant protection convention},
  http://www.fao.org/plant-health-2020/about/en.

\bibitem{fasoula2012nonstop}
Dionysia~A Fasoula, \emph{Nonstop selection for high and stable crop yield by
  two prognostic equations to reduce yield losses}, Agriculture \textbf{2}
  (2012), no.~3, 211--227.

\bibitem{gao2020framework}
Demin Gao, Quan Sun, Bin Hu, and Shuo Zhang, \emph{A framework for agricultural
  pest and disease monitoring based on internet-of-things and unmanned aerial
  vehicles}, Sensors \textbf{20} (2020), no.~5, 1487.

\bibitem{goodfellow2020generative}
Ian Goodfellow, Jean Pouget-Abadie, Mehdi Mirza, Bing Xu, David Warde-Farley,
  Sherjil Ozair, Aaron Courville, and Yoshua Bengio, \emph{Generative
  adversarial networks}, Communications of the ACM \textbf{63} (2020), no.~11,
  139--144.

\bibitem{hughes2015open}
David Hughes, Marcel Salath{\'e}, et~al., \emph{An open access repository of
  images on plant health to enable the development of mobile disease
  diagnostics}, arXiv preprint arXiv:1511.08060 (2015).

\bibitem{joshic2020}
C.~Joshi, \emph{Generative adversarial networks (gans) for synthetic dataset
  generation with binary classes},
  https://datasciencecampus.ons.gov.uk/projects/generative-adversarial-networks-gans-for-synthetic-dataset-generation-with-binary-classes/
  (2020).

\bibitem{kingma2013auto}
Diederik~P Kingma and Max Welling, \emph{Auto-encoding variational bayes},
  arXiv preprint arXiv:1312.6114 (2013).

\bibitem{mirza2014conditional}
Mehdi Mirza and Simon Osindero, \emph{Conditional generative adversarial nets},
  arXiv preprint arXiv:1411.1784 (2014).

\bibitem{nazki2020unsupervised}
Haseeb Nazki, Sook Yoon, Alvaro Fuentes, and Dong~Sun Park, \emph{Unsupervised
  image translation using adversarial networks for improved plant disease
  recognition}, Computers and Electronics in Agriculture \textbf{168} (2020),
  105117.

\bibitem{perez2017effectiveness}
Luis Perez and Jason Wang, \emph{The effectiveness of data augmentation in
  image classification using deep learning}, arXiv preprint arXiv:1712.04621
  (2017).

\bibitem{radford2015unsupervised}
Alec Radford, Luke Metz, and Soumith Chintala, \emph{Unsupervised
  representation learning with deep convolutional generative adversarial
  networks}, arXiv preprint arXiv:1511.06434 (2015).

\bibitem{simonyan2014very}
Karen Simonyan and Andrew Zisserman, \emph{Very deep convolutional networks for
  large-scale image recognition}, arXiv preprint arXiv:1409.1556 (2014).

\bibitem{singh2012taro}
Davinder Singh, Grahame Jackson, Danny Hunter, Robert Fullerton, Vincent Lebot,
  Mary Taylor, Tolo Iosefa, Tom Okpul, and Joy Tyson, \emph{Taro leaf
  blight—a threat to food security}, Agriculture \textbf{2} (2012), no.~3,
  182--203.

\bibitem{sugiura2016field}
Ryo Sugiura, Shogo Tsuda, Seiji Tamiya, Atsushi Itoh, Kentaro Nishiwaki,
  Noriyuki Murakami, Yukinori Shibuya, Masayuki Hirafuji, and Stephen Nuske,
  \emph{Field phenotyping system for the assessment of potato late blight
  resistance using rgb imagery from an unmanned aerial vehicle}, Biosystems
  engineering \textbf{148} (2016), 1--10.

\bibitem{thapa2020plant}
Ranjita Thapa, Noah Snavely, Serge Belongie, and Awais Khan, \emph{The plant
  pathology 2020 challenge dataset to classify foliar disease of apples}, arXiv
  preprint arXiv:2004.11958 (2020).

\bibitem{wang2017automatic}
Guan Wang, Yu~Sun, and Jianxin Wang, \emph{Automatic image-based plant disease
  severity estimation using deep learning}, Computational intelligence and
  neuroscience \textbf{2017} (2017).

\bibitem{rwightman}
R.~Wightman, \emph{Genetic efficient nets for pytorchs},
  https://github.com/rwightman/efficientdet-pytorch (2021).

\end{thebibliography}

\end{document}